\documentclass[conference]{IEEEtran}
\IEEEoverridecommandlockouts
\usepackage{cite}
\usepackage{amsmath,amssymb,amsfonts}
\usepackage{algorithmic}
\usepackage{graphicx}
\usepackage{tabularx}
\usepackage{hyperref}
\usepackage[ruled, linesnumbered, vlined]
{algorithm2e}
\hypersetup{
    colorlinks=true,
    linkcolor=blue,  % You can set the color to your preference
    linktoc=page
}
\usepackage{graphicx}
\usepackage{textcomp}
\usepackage{xcolor}
\def\BibTeX{{\rm B\kern-.05em{\sc i\kern-.025em b}\kern-.08em
    T\kern-.1667em\lower.7ex\hbox{E}\kern-.125emX}}
\begin{document}

\title{Color Recognition in Challenging Lighting Environments: CNN Approach\\
\thanks{Identify applicable funding agency here. If none, delete this.}
}

\author{\IEEEauthorblockN{1\textsuperscript{st} \textbf{Nizamuddin Maitlo$^*$}}
\IEEEauthorblockA{\textit{Department of Computer Science} \\
\textit{Sukkur IBA University}\\
Sukkur, Pakistan \\
Corresponding Author: nizamuddin.cs@iba-suk.edu.pk
}
\and
\IEEEauthorblockN{2\textsuperscript{nd} Nooruddin Noonari}
\IEEEauthorblockA{\textit{Department of Computer Science} \\
\textit{IBA Community College, Naushahro Feroze}\\
Sindh, Pakistan\\
Corresponding Author: noor.cs2@yahoo.com}

\and
\IEEEauthorblockN{3\textsuperscript{rd} Sajid Ahmed Ghanghro}
\IEEEauthorblockA{\textit{Depeartment of Computer Science
} \\
\textit{Saitama University, Japan}\\
\textit{S.A.LU, Khairpur, Sindh, Pakistan}\\
ahmed.s.285@ms.saitama-u.ac.jp\\
sajid.ghanghro@salu.edu.pk}

\and
\IEEEauthorblockN{4\textsuperscript{th} Sathishkumar Duraisamy}
\IEEEauthorblockA{\textit{Department of Mechanical Engineering} \\
\textit{Kathir College of Engineering}\\
Tamilnadu, India.\\
sathizkumard@gmail.com}

\and
\IEEEauthorblockN{5\textsuperscript{th} Fayaz Ahmed}
\IEEEauthorblockA{\textit{Institute of Computer Science} \\
\textit{S.A.LU, Khairpur}\\
Sindh, Pakistan\\
liaquatfayaz0@gmail.com}

}

\maketitle

\begin{abstract}
Light plays a vital role in vision either human or machine vision, the perceived color is always based on the lighting conditions of the surroundings. Researchers are working to enhance the color detection techniques for the application of computer vision. They have implemented proposed several methods using different color detection approaches but still, there is a gap that can be filled.

To address this issue, a color detection method, which is based on a Convolutional Neural Network (CNN), is proposed. Firstly, image segmentation is performed using the edge detection segmentation technique to specify the object and then the segmented object is fed to the Convolutional Neural Network trained to detect the color of an object in different lighting conditions. It is experimentally verified that our method can substantially enhance the robustness of color detection in different lighting conditions, and our method performed better results than existing methods.

\end{abstract}


\section{Introduction}
Recently researchers have been working on different color detection techniques to enhance the computer's vision applications to be more accurate and precise. In computer vision, color plays a very important role in different techniques such as object detection, object tracking, image retrieval, object recognition, and so on. While comparing color images with gray-scale images color images provide more information than gray-scale images. \cite{b1}.

Several techniques have been implemented by researchers by using color detection such as \cite{b2} a Convolutional Neural Network is implemented to detect vehicle color. Further, they mentioned that the CNNs are naturally designed to perform classification based on the shape of the vehicle but their research proved that CNNs can be trained to perform color classification and can perform very well. In their method they converted the input image into different two color spaces, HSV and CIE Lab, and performed them to some CNN architecture. In this paper \cite{b3}, researchers explored an object detection technique based on fusion architecture. Two automotive datasets namely the KAIST multispectral pedestrian dataset and FLIR thermal object detection dataset are explored for both these sensors. In their research, they found that both models are best in some aspects and outperformed in different conditions. According to their complementary nature, the Color model outperforms Thermal in day conditions and the Thermal model outperforms color in night conditions. They have used CNN fusion architecture of mid-level that performed significantly better than the existing baseline models. They have observed 0.6\%  improvements in experimental results compared to existing baseline methods. In \cite{b4}, researchers investigated the impact of basic image processing and contrast enhancement techniques such as CLAHE and MSRCR on CNN-based skin cancer detection. The findings indicate that, in comparison to MSRCR, CLAHE is preferable for color image enhancement in early skin cancer detection using CNN. Nevertheless, both the original and CLAHE-enhanced datasets demonstrated equivalent accuracy during training and validation. The primary finding of this study suggests that image contrast enhancement is unnecessary for effective skin cancer screening. In this study \cite{b5}, researchers have introduced a light field imaging technique designed to address challenges in underwater imaging under low-light conditions. Their approach focuses on mitigating the issue of light scattering in light field images through the utilization of deep convolutional neural networks incorporating depth estimation. Additionally, They employ a color correction method based on spectral characteristics to restore diminished colors. The experimental outcomes demonstrate the efficacy of their proposed method in addressing real-world challenges associated with underwater imaging.

To address the issue of color detection, numerous existing methods have demonstrated the capability to accurately identify a wide range of colors. Nonetheless, this task remains challenging due to the influence of environmental factors. Achieving precise color recognition is crucial for applications involving visual inspection. One well-known approach for color detection is the utilization of the HSV (Hue, Saturation, and Value) color space, as documented in previous works \cite{b6}\cite{b7}. The effectiveness of this method greatly depends on the accurate determination of the minimum and maximum range for a specific color, a process often guided by statistical techniques. Failure to do so can result in the susceptibility of the specified color range to environmental variations, including changes in lighting conditions. HSV color space is particularly robust when it comes to detecting a specific color with precision. In such cases, the range can be carefully calculated to ensure accurate color detection. However, the HSV color detection method may not be universally applicable, especially in scenarios where the objective is to detect multiple colors, and some of these colors exhibit similar properties. Therefore, there is a pressing need for the development of innovative methods that can emulate the human visual system's ability to accurately perceive colors. This is essential because color perception can be significantly influenced by various sources of illumination, such as natural sunlight, LED lighting, and CCFL (Cold Cathode Fluorescent Lamp) lighting.

In response to the aforementioned challenge, we introduce a novel color detection approach that leverages the power of Convolutional Neural Networks (CNNs). Our method combines image segmentation with edge detection techniques to accurately delineate the object of interest. Subsequently, this segmented object is input into a trained Convolutional Neural Network, specifically designed to excel in detecting object colors across varying lighting conditions. Experimental validation demonstrates the remarkable effectiveness of our approach in enhancing color detection robustness under diverse lighting scenarios, surpassing the performance of established methods.

The subsequent sections of this article are structured as follows:
\begin{itemize}

\item Literature Review (Section II): This section provides a concise overview of the relevant literature in the field.

\item Proposed Methodology for the Indoor Navigation System of Autonomous Vehicles (Section III): Section III offers a brief description of the methodology proposed for the indoor navigation system of autonomous vehicles.
\item Results and Discussion (Section IV): In this section, you will find the results and a detailed discussion of the outcomes obtained through the application of the proposed methods.

\item Conclusion (Section V): The article concludes in this section, summarizing the key findings and contributions of the study.
\end{itemize}

This organization provides a clear structure for readers to understand the flow of the article and easily locate specific information within it.

\section{Color Spectrum System}

\begin{figure}[htbp]
\centering
\includegraphics[width=0.5\textwidth]{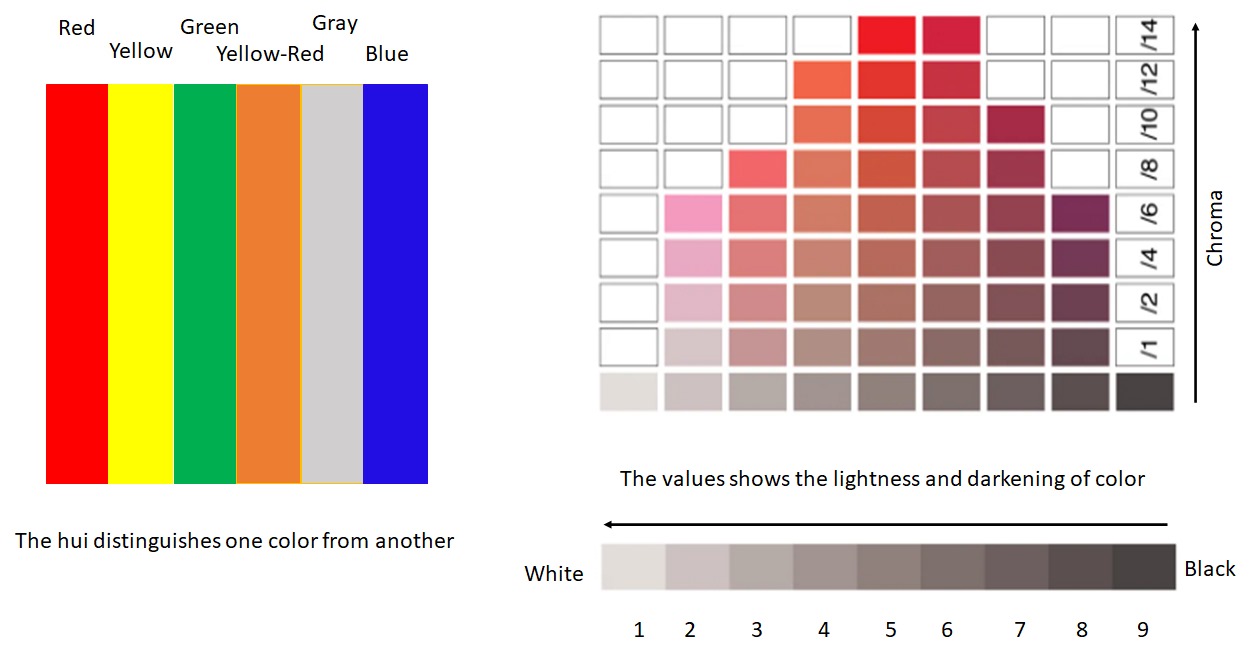} % Adjust the width as needed
\caption{Spectrum of six different colors, whose brightness changes from dark to light.}
\label{pic01}
\end{figure}

In \autoref{pic01} spectrum of different colors, whose brightness changes from dark to light is shown in the image. Our algorithm is focused on this change in the lighting of color and will enhance the color detection capability of the applications.

\section{Literature Review}

Currently, researchers are exploring various ways to improve the effectiveness and efficiency of color detection by proposing new techniques, specifically in different lighting conditions. Color is very useful in several applications such as object detection, edge detection, and so on but color detection through baseline techniques is still limited. These methods are as follows.

This study \cite{b8} focuses on developing a high-performance traffic sign detection system using deep learning techniques. The study delves into two key aspects of the recognition system: the color space used to process the input video and the structure of the deep learning network. The study combines three different types of network models based on faster R-CNN and R-FCN architectures and thoroughly evaluates six different color spaces, namely RGB, Normalized RGB, Ruta's RYG, YCbCr, HSV, and CIE Lab. The experiment consists of end-to-end testing on a traffic light dataset with a resolution of 1280×720. Notably, simulations show that the best performance is achieved when the RGB color space and the Faster R-CNN model are used together.

This paper \cite{b9} uses fine-grained convolutional neural networks (CNNs) to accurately predict scene lighting, avoiding the use of hand-crafted features prevalent in previous methods. CNNs operate exclusively in the spatial domain and take image patches as input, and their architecture consists of one convolutional layer (including max pooling), a fully connected layer, and three dedicated output nodes. It is worth noting that this network architecture seamlessly integrates feature learning and regression into a single optimization process, thus creating a more effective model for scene illumination estimation. This approach demonstrates excellent performance and sets a new standard for standard RAW image datasets. Experimental results highlight the effectiveness of the proposed CNN in estimating scene illumination. Notably, preliminary experiments on images with spatially varying illumination characteristics confirmed the robustness and stability of the local illumination estimation capabilities of the CNN.

This paper \cite{b10} proposes a new low-light image enhancement model called MSR-net, which combines the principles of Retinex theory and convolutional neural networks (CNNs). By establishing an equivalence between multiscale Retinex and feedforward CNN, the model directly learns the global mapping between dark and light images, reformulating low-light image enhancement as a machine-learning problem. Compared with traditional methods, most MSR network parameters are optimized through error backpropagation, eliminating the dependence on manual parameter tuning. Experimental evaluation of complex images shows that the model qualitatively and quantitatively outperforms state-of-the-art methods, highlighting its potential to advance the field of low-light image enhancement.

This study \cite {b11} developed a robust crack detection method by implementing transfer learning as an alternative to training original neural networks. The study explores three deep learning approaches: 1) building an accurate convolutional neural network from scratch, 2) using the output functions of the VGG16 network architecture pre-trained on the ImageNet dataset, and 3) fine-tuning the top layer of VGG16. To mitigate problems associated with limited and imbalanced training datasets, data augmentation is used. The dataset contains fatigue test photographs and inspection images taken under different conditions such as uncontrolled distance, illumination, angle and blur. It is worth noting that the increase in data significantly helped improve the accuracy index, which increased the performance of the three verification methods by 5\%, 2\%, and 5\%, respectively. This comprehensive approach to crack detection highlights the effectiveness of transfer learning in addressing real-world variations and limitations in training datasets.

This paper \cite{b12} proposes a convolutional neural network (CNN)-based model that enables end-to-end real-time lighting estimation in mixed reality environments even without a priori scene information. Ambient lighting is represented using spherical harmonics (SH), specifically designed to effectively capture area lighting. The proposed CNN architecture takes RGB images as input and quickly recognizes ambient lighting. This study differs from traditional CNN-based illumination estimation methods, which require a highly optimized deep neural network architecture with a small number of parameters. This design enables the model to skillfully learn complex lighting scenarios from realistic high dynamic range (HDR) environment images, highlighting its potential for robust and efficient lighting estimation in a variety of mixed reality environments.

 The study \cite {b13} presents a system designed for face recognition in low-light conditions, using a dual strategy of low-light image enhancement and face detection. Another study \cite{b14} focused on precise modeling of object color recognition thresholds under different lighting conditions. In addition, a separate study proposed a new data-driven model \cite{b15} specifically designed to estimate high dynamic range (HDR) lighting conditions from a single low dynamic range (LDR) spherical panoramic image. Notably, the inherently complex and flawed task of illumination estimation is further complicated by the lack of reliable illumination data, which poses a major challenge for the practical application of data-driven methods. Subsequent research has investigated object detection in complex environments. One study \cite{b16} calls these situations complex environments and points to advances in deep learning-based object detection. Another study \cite{b17} pioneered the use of high dynamic range (HDR) imaging in object detection, increasing the ability to capture and process a wide dynamic range of a scene, similar to the human eye. Other studies \cite{b18},\cite{b19} improve low-light images by implementing learning techniques, with applications ranging from light-field image enhancement to low-light image enhancement. Finally, another paper \cite{b20} shifts the focus to using image enhancement techniques to improve object detection performance rather than improving the quality of human sensory perception.

\section{Methodology}
The concept of employing color spectrums for training neural networks drew inspiration from nature. In numerous instances, a color's appearance can be noticeably influenced by environmental conditions, causing subtle shifts in its visual perception. For instance, a transition from light yellow to a slightly darker shade of yellow might still be perceived as yellow by the human eye. However, in the context of machine vision, this issue becomes considerably more consequential. Accurate color measurement becomes imperative in certain applications to effectively differentiate between two similar objects based on their color characteristics.
\subsection{Dataset}

The proposed CNN is trained using a dataset of 250 images of different variations of colors in different lighting conditions. The dataset is divided into training and testing for training and testing purposes, we have used 200 images for training and the remaining 50 for the testing purpose of the model.

\subsection{Experimental Settings}
To perform the experiments we have used the system with CPU: Intel Core i7 8700K; memory: 16 GB RAM; OS: Ubuntu 16.04 for training and testing the proposed framework with different lighting conditions.

\subsection{CNN}
We have proposed a Convolutional Neural Network (CNN) architecture for color classification containing three convolutional layers, along with Rectified Linear Unit (ReLU) activation, augmenting feature extraction, and infusing non-linearity. The first three convolutional layers are conjoined with Max Pooling operations to effectuate spatial dimension reduction. Subsequent to this convolutional sequence, three fully connected layers perform feature abstraction, ultimately feeding into an output layer tailored for color categorization.
\begin{figure}[htbp]
\centering
\includegraphics[width=0.5\textwidth]{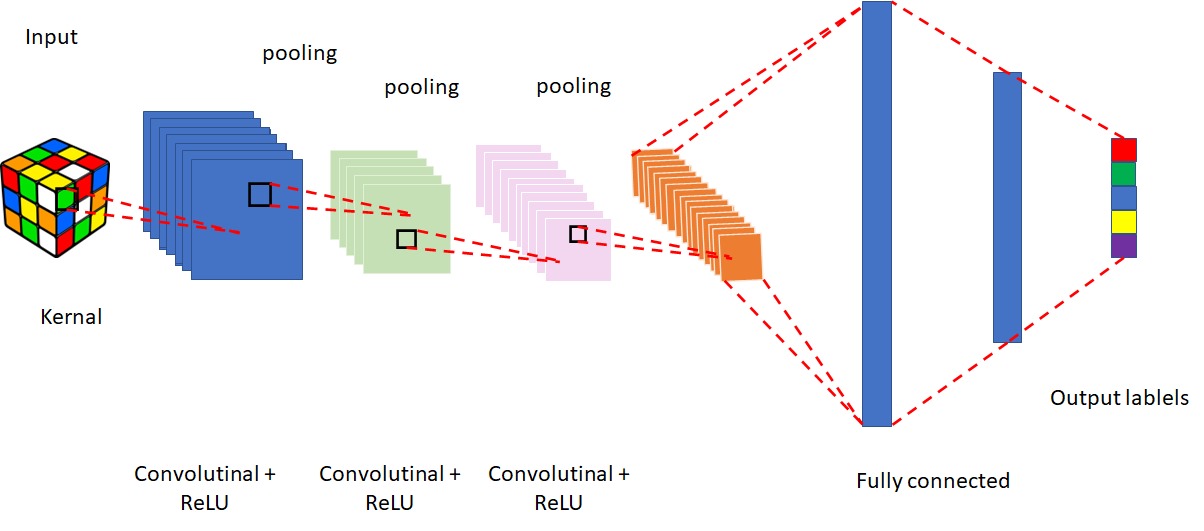} % Adjust the width as needed
\caption{Spectrum of six different colors, whose brightness changes from dark to light.}
\label{pic02}
\end{figure}

A typical CNN layer can be represented with the following equations:

\begin{equation}
    z_i = (w * x)_i + b_i
\end{equation}

\begin{equation}
    a_i = f(z_i)
\end{equation}

Here, $z_i$ is the weighted sum of the input activations, $w$ is the weight matrix, $x$ is the input feature map, $b_i$ is the bias term, and $f(\cdot)$ is the activation function.

The convolution operation in equation (1) is defined as:

\begin{equation}
    (w * x)_i = \sum_{m=0}^{M-1} \sum_{n=0}^{N-1} w_{m,n} \cdot x_{i+m, n}
\end{equation}

where $w_{m,n}$ is the weight at position $(m, n)$ in the filter, and $x_{i+m, n}$ is the input at position $(i+m, n)$.

The activation function $f(z_i)$ can be a non-linear function such as the Rectified Linear Unit (ReLU):

\begin{equation}
    f(z_i) = \max(0, z_i)
\end{equation}

\subsubsection{Pooling Layer}

A pooling layer can be represented as follows:

\begin{equation}
    y_{i,j} = \max_{m=0}^{K-1} \max_{n=0}^{K-1} x_{i \cdot K + m, j \cdot K + n}
\end{equation}

Here, $y_{i,j}$ is the output of the pooling layer, and $K$ is the pooling size.

\subsection{Proposed Framework}
The concept of employing color spectrums in training neural networks draws inspiration from nature. In various instances, environmental conditions can notably influence a color, altering its visual appearance. Consider, for instance, the transition from light yellow to a slightly darker shade of yellow; to the human eye, it remains categorized as yellow. While this might not be a significant concern in human vision, it holds crucial importance in machine vision. Achieving precise color measurements becomes imperative in certain applications where distinguishing between two similar objects relies on their color properties. 

We have proposed a framework in which the first image is fed to the region proposal module which detects the object using Sobel edge detection and then edges are linked to form the object. Further, the detected objects are merged and sent to CNN for color detection to the CNN Training and classification module in which first the model is trained using a color dataset and then the model classifies the color based on the training.

\begin{figure}[htbp]
\centering
\includegraphics[width=0.5\textwidth]{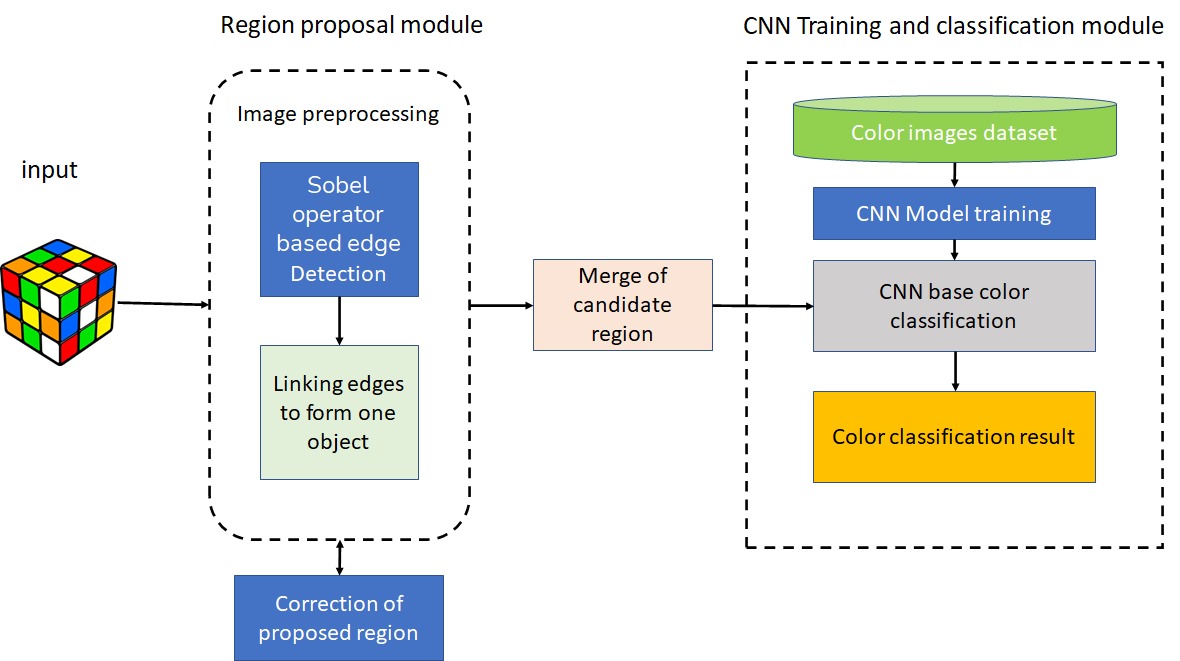} % Adjust the width as needed
\caption{Spectrum of six different colors, whose brightness changes from dark to light.}
\label{pic03}
\end{figure}

\subsection{Bounding Box Overview}
A bounding box is a rectangular frame that encapsulates and precisely outlines the spatial extent of an object or region within an image. Widely used in computer vision tasks, particularly object detection, bounding boxes provide a standardized representation, specifying the object's location and facilitating subsequent analyses. These boxes are defined by their coordinates, typically represented by the top-left and bottom-right corners, enabling efficient localization and recognition of objects in image datasets.

Our approach entails a systematic process of object detection, wherein bounding boxes are employed as integral annotations. Before the subsequent identification of color attributes within the designated regions or objects, the establishment of bounding boxes is imperative. This preliminary step in the computational framework serves to precisely localize and delineate objects of interest within the visual domain. By employing this methodological sequence, we aim to enhance the accuracy and specificity of color identification processes, thus contributing to the robustness and efficacy of our overarching scientific inquiry in computer vision applications.

\subsection{Pusodocode}

The provided pseudocode outlines a method for extracting color cubes from a given bounding box in an RGB image. The nested loops iterate over a 3x3 grid within the bounding box, determining the center coordinates for each color cube. Subsequently, a sequence of color cubes is extracted from the bounding box image based on a specified size. These color cubes are then input into a neural network model, denoted as RCCNet, for predictive analysis. The algorithm facilitates the systematic processing of localized regions within the bounding box, enabling the recognition or extraction of features relevant to the specified task, such as color identification.

\begin{algorithm}
\caption{Process Bounding Box}
\SetKwInOut{Input}{Input}
\SetKwInOut{Output}{Output}
\Input{$x, y, w, h$ // Bounding box information by MBR, $img$ // original image RGB, $size=32$ // Size for each cube to recognize}
\Output{$rectCubeImg$ // sequence of color cubes, $boundBoxImg = img(Range(x, x+w), Range(y, y+h))$}

\For{$i = 0$ \KwTo $2$}{
    $x1 = \frac{w}{3} \cdot \frac{1}{2} + i \cdot 3$\;
    \For{$j = 0$ \KwTo $2$}{
        $y1 = \frac{h}{3} \cdot \frac{1}{2} + j \cdot y$\;
        $rectCubeImg = boundBoxImg(Range(x1 - \frac{size}{2}, x1 + \frac{size}{2}), Range(y1 - \frac{size}{2}, y1 + \frac{size}{2}))$\;
        RCCNet.Predict($rectCubeImg$)\;
    }
}
\end{algorithm}

\section{Experiment and Results}

\begin{figure}[htbp]
\centering
\includegraphics[width=0.4\textwidth]{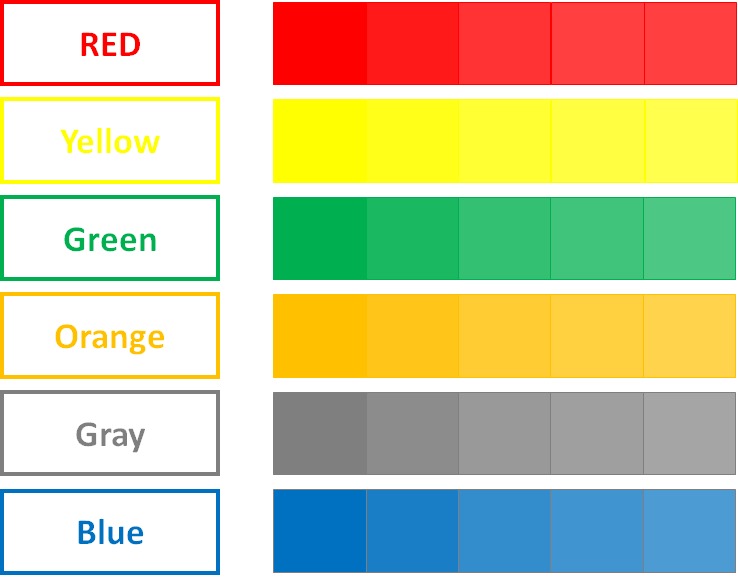} % Adjust the width as needed
\caption{Spectrum of six different colors, whose brightness changes from dark to light.}
\label{pic04}
\end{figure}

\begin{figure}[htbp]
\centering
\includegraphics[width=0.5\textwidth]{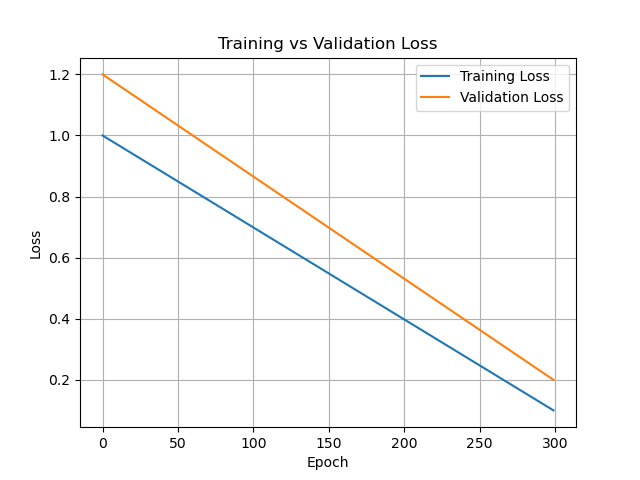} % Adjust the width as needed
\caption{Training loss and Validation Loss}
\label{pic05}
\end{figure}

The graph denoted as \autoref{pic05} illustrates the trajectory of training and validation loss. Notably, the observed pattern indicates a linear decrease in both training and validation loss as the number of epochs progresses.
 
\begin{figure}[htbp]
\centering
\includegraphics[width=0.5\textwidth]{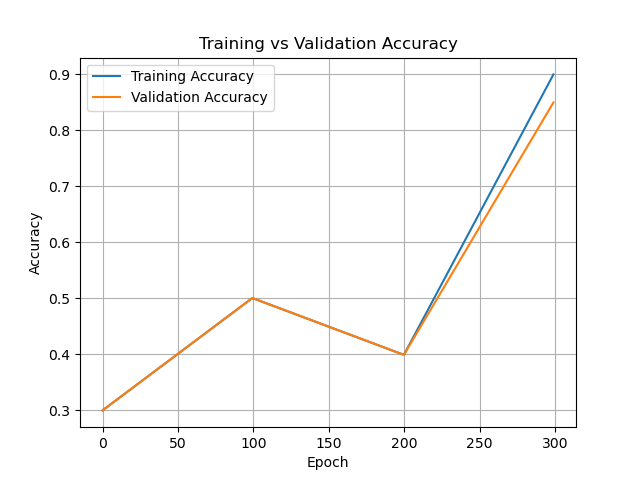} % Adjust the width as needed
\caption{Training Accuracy and Validation Accuracy}
\label{pic06}
\end{figure}

Displayed in \autoref{pic06} are the training and validation accuracy curves. This representation reveals that the model's accuracy experiences an initial ascent from the first epoch, followed by a marginal decline between epochs 100 and 200. Subsequently, beyond the 200th epoch, the accuracy graphs demonstrate a linear upward trend.

\begin{figure}[htbp]
\centering
\includegraphics[width=0.5\textwidth]{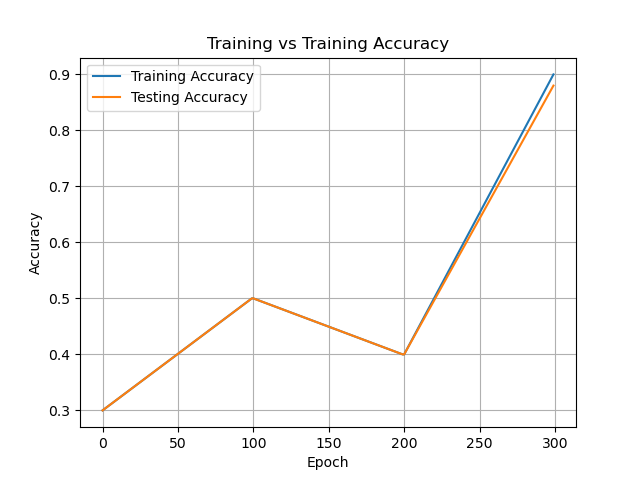} % Adjust the width as needed
\caption{Training Accuracy and Testing Accuracy}
\label{pic07}
\end{figure}

Similarly, the graph referenced as \autoref{pic07} delineates the training and testing accuracy. Analogous to the pattern observed in \autoref{pic06}, the accuracy initially rises from the onset and undergoes a minor dip between epochs 100 and 200, only to resume a linear increase thereafter.

\begin{figure}[htbp]
\centering
\includegraphics[width=0.5\textwidth]{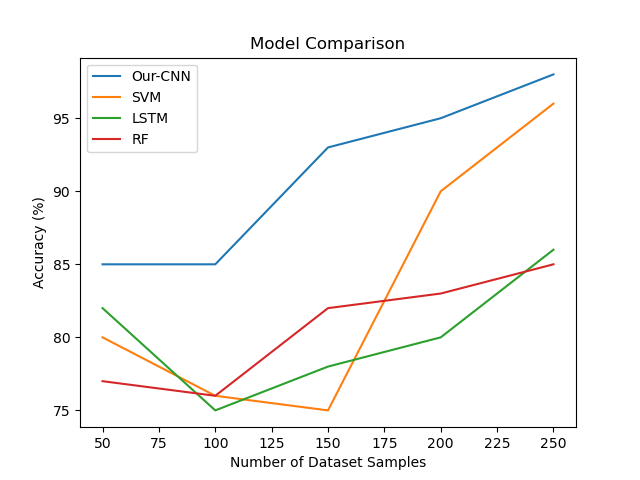} % Adjust the width as needed
\caption{Model Comparison with other approaches over different No of Images}
\label{pic08}
\end{figure}

The graph in figure \autoref{pic08} shows that our proposed Convolutional Neural Network (CNN) surpasses Support Vector Machines (SVM), Long Short-Term Memory (LSTM), and Random Forest (RF) in performance. The CNN's ability to automatically learn intricate features from raw data, capture spatial hierarchies in image data, and engage in end-to-end learning distinguishes it, resulting in superior outcomes compared to traditional models.

\section{Our Results}
\begin{figure}[htbp]
\centering
\includegraphics[width=0.5\textwidth]{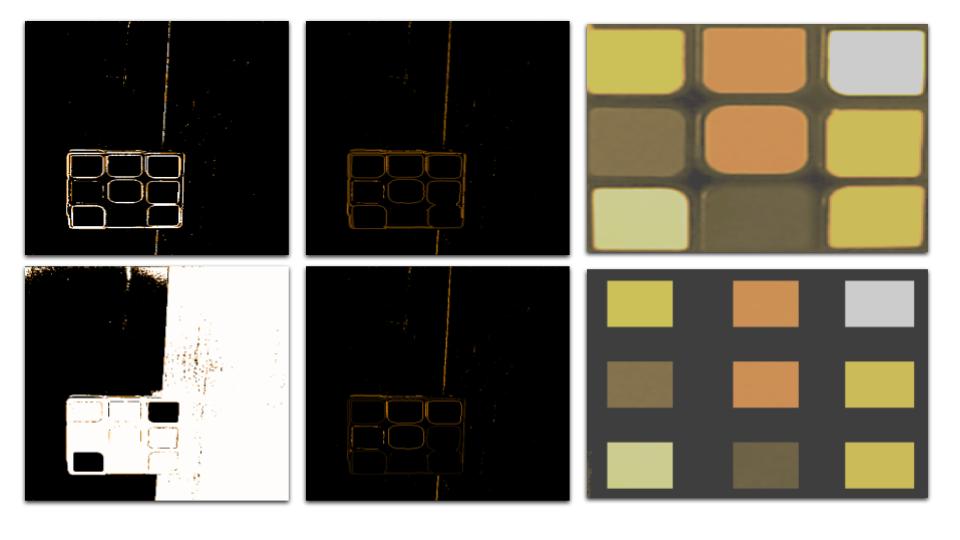} % Adjust the width as needed
\caption{Our Purposed Method Results}
\label{pic09}
\end{figure}

The image capture process is initiated using an imaging device, followed by preprocessing involving Gaussian Blur. Subsequent to this, the Adaptive Threshold method surpasses the standard threshold function in effectively segmenting the bounding box, as depicted in \autoref{pic09}. Employing the contour method on the output of the adaptive threshold facilitates the detection of the bounding box, with further processing of identified contours to ascertain the bounding box's edges. Remarkably, the largest contour in the image is designated as the bounding box, highlighted by the outer red pixels in \autoref{pic09}. The Minimum Bounding Rectangle (MBR) method is then applied to extract crucial details such as position, width, and height for subsequent cropping from the input RGB image. Following the acquisition of bounding box specifics via the MBR method, a feature extraction procedure is initiated to isolate the bounding box from the RGB image. Consequently, color features are extracted from the bounding box to facilitate recognition. The results of the bounding box detection and feature extraction are visually represented in \autoref{pic09}. Ultimately, our Convolutional Neural Network (CNN) is employed for the recognition of each color within the detected bounding box. Significantly, CNN's training encompasses variations in illuminations, ensuring accurate color recognition based on distinctive spectral characteristics.

\section{Discussion}
In response to the challenge of robust color detection in diverse lighting conditions, we propose a Convolutional Neural Network (CNN)-based method. Initially, image segmentation is conducted using edge detection segmentation to precisely identify the object. Subsequently, the segmented object is input into a CNN trained for color detection under varying lighting conditions. Experimental validation demonstrates that our approach significantly improves the robustness of color detection across diverse lighting scenarios, surpassing the performance of existing methods.

Our study underscores the superior performance of our proposed CNN in contrast to Support Vector Machines (SVM), Long Short-Term Memory (LSTM), and Random Forest (RF). The CNN's inherent capacity to autonomously extract intricate features from raw data, capture spatial hierarchies within image data, and facilitate end-to-end learning sets it apart, yielding outcomes superior to traditional models. This research contributes valuable insights into the efficacy of CNNs for enhanced color detection, particularly in challenging environmental conditions.

\section{Conclusion}

Light plays a crucial role in both human and machine vision, where the perceived color is inherently influenced by the ambient lighting conditions. To advance color detection techniques for computer vision applications, researchers have explored various methods using distinct color detection approaches. However, despite these efforts, a discernible gap remains in the existing literature.

In response to this gap, we propose a color detection methodology based on a Convolutional Neural Network (CNN). Initially, image segmentation is conducted using an edge detection segmentation technique to delineate the object of interest. Subsequently, the segmented object is input into a Convolutional Neural Network that has been trained to identify the color of objects under diverse lighting conditions. Through experimental validation, our method demonstrates a notable improvement in the robustness of color detection across varying lighting conditions. Our approach outperforms existing methods, affirming its efficacy in addressing the challenges associated with color detection in diverse lighting environments.

\vspace{12pt}
\color{red}
IEEE conference templates contain guidance text for composing and formatting conference papers. Please ensure that all template text is removed from your conference paper prior to submission to the conference. Failure to remove the template text from your paper may result in your paper not being published.


\begin{thebibliography}{00}
\bibitem{b1} Deng, S., Tian, Y., Hu, X., Wei, P., \& Qin, M. (2012). Application of new advanced CNN structure with adaptive thresholds to color edge detection. Communications in Nonlinear Science and Numerical Simulation, 17(4), 1637-1648.
\bibitem{b2} Rachmadi, R. F., \& Purnama, I. (2015). Vehicle color recognition using convolutional neural network. arXiv preprint arXiv:1510.07391.
\bibitem{b3} Yadav, R., Samir, A., Rashed, H., Yogamani, S., \& Dahyot, R. (2020). Cnn based color and thermal image fusion for object detection in automated driving. Irish Machine Vision and Image Processing.
\bibitem{b4} Setiawan, A. W. (2020, February). Effect of color enhancement on early detection of skin cancer using convolutional neural network. In 2020 IEEE International Conference on Informatics, IoT, and Enabling Technologies (ICIoT) (pp. 100-103). IEEE.
\bibitem{b5} Lu, H., Li, Y., Uemura, T., Kim, H., \ Serikawa, S. (2018). Low illumination underwater light field images reconstruction using deep convolutional neural networks. Future Generation Computer Systems, 82, 142-148.
\bibitem{b6} Bahroun, S. (2023). Combining Color Texture and Shape Features In a Multi-Input Convolutional Neural Network for Efficient Face Recognition In Unconstrained Environments.
\bibitem{b7}  Pei, X., hong Zhao, Y., Chen, L., Guo, Q., Duan, Z., Pan, Y., \& Hou, H. (2023). Robustness of machine learning to color, size change, normalization, and image enhancement on micrograph datasets with large sample differences. Materials \& Design, 232, 112086.
\bibitem{b8} Kim, H. K., Park, J. H., \& Jung, H. Y. (2018). An efficient color space for deep-learning based traffic light recognition. Journal of Advanced Transportation, 2018, 1-12.
\bibitem{b9} Bianco, S., Cusano, C., \& Schettini, R. (2015). Color constancy using CNNs. In Proceedings of the IEEE conference on computer vision and pattern recognition workshops (pp. 81-89).
\bibitem{b10} Shen, L., Yue, Z., Feng, F., Chen, Q., Liu, S., \& Ma, J. (2017). Msr-net: Low-light image enhancement using deep convolutional network. arXiv preprint arXiv:1711.02488.
\bibitem{b11}Dung, C. V., Sekiya, H., Hirano, S., Okatani, T., \& Miki, C. (2019). A vision-based method for crack detection in gusset plate welded joints of steel bridges using deep convolutional neural networks. Automation in Construction, 102, 217-229.
\bibitem{b12} Marques, B. A. D., Clua, E. W. G., Montenegro, A. A., \& Vasconcelos, C. N. (2022). Spatially and color consistent environment lighting estimation using deep neural networks for mixed reality. Computers \& Graphics, 102, 257-268.
\bibitem{b13} Le, C., \& Mohd, T. K. (2022, June). Facial Detection in Low Light Environments Using OpenCV. In 2022 IEEE World AI IoT Congress (AIIoT) (pp. 624-628). IEEE.
\bibitem{b14} Ponting, S., Morimoto, T., \& Smithson, H. (2018). Modelling surface color discrimination under different lighting environments using. Journal of the Optical Society of America A, 35(4), B256-B266.
\bibitem{b15} Gkitsas, V., Zioulis, N., Alvarez, F., Zarpalas, D., \& Daras, P. (2020). Deep lighting environment map estimation from spherical panoramas. In Proceedings of the IEEE/CVF Conference on Computer Vision and Pattern Recognition Workshops (pp. 640-641).
\bibitem{b16} Ahmed, M., Hashmi, K. A., Pagani, A., Liwicki, M., Stricker, D., \& Afzal, M. Z. (2021). Survey and performance analysis of deep learning based object detection in challenging environments. Sensors, 21(15), 5116.
\bibitem{b17} Mukherjee, R., Bessa, M., Melo-Pinto, P., \& Chalmers, A. (2021). Object detection under challenging lighting conditions using high dynamic range imagery. IEEE Access, 9, 77771-77783.
\bibitem{b18} Ge, Z., Song, L., \& Lam, E. Y. (2020, November). Light field image restoration in low-light environment. In SPIE Future Sensing Technologies (Vol. 11525, pp. 300-305). SPIE.
\bibitem{b19} Li, C., Guo, J., Porikli, F., \& Pang, Y. (2018). LightenNet: A convolutional neural network for weakly illuminated image enhancement. Pattern recognition letters, 104, 15-22.
\bibitem{b20} Guo, H., Lu, T., \& Wu, Y. (2021, January). Dynamic low-light image enhancement for object detection via end-to-end training. In 2020 25th International Conference on Pattern Recognition (ICPR) (pp. 5611-5618). IEEE.




\bibitem{b21} R. S. Run and Z. Y. Xiao, “Indoor Autonomous Vehicle Navigation—A Feasibility Study Based on Infrared Technology,” Appl. Syst. Innov. 1(1), 4–4 (2018).
\bibitem{b22} R. Nock and F. Nielsen, “Statistical Region Merging,” IEEE Trans. Pattern Anal. Mach. Intell. 26(11), 1452–1458 (2004).
\bibitem{b23} H. D. Cheng, X. H. Jiang, Y. Sun and J. Wang, “Color image segmentation: Advances and prospects,” Pattern Recognit. 34(12), 2259-2281 (2001).
\bibitem{b24} S. Suzuki and K. Abe, “Topological Structural Analysis of Digitized Binary Images by Border Following,” Comput. Vis. Graph. Image Process 30(1), 32-46 (1985).
\bibitem{b25} J. Flusser and T. Suk, “Rotation Moment Invariants for Recognition of Symmetric Objects,” IEEE Trans. Image Process. 15(12), 3784–3790 (2006).
\bibliographystyle{alpha}
\bibliography{sample}


\end{thebibliography}
\end{document}